\documentclass[preprint, 10pt]{elsarticle}
\usepackage[margin=2cm]{geometry}

\usepackage[utf8]{inputenc}

\usepackage{multirow}
\usepackage{tabu}
\usepackage{algorithm}
\usepackage{algorithmic}
\usepackage{graphicx}
\usepackage{amsmath}

\makeatletter
\def\ps@pprintTitle{%
   \let\@oddhead\@empty
   \let\@evenhead\@empty
   \let\@oddfoot\@empty
   \let\@evenfoot\@oddfoot
}
\makeatother

\begin{document}

\begin{frontmatter}


\title{Deep Learning for Classification and Severity Estimation of Coffee Leaf Biotic Stress}


\author[a,b]{Jos{\'e} G. M. Esgario\corref{correspondingauthor}}
\cortext[correspondingauthor]{Corresponding author}
\ead{guilherme.esgario@gmail.com}

\author[a,b,c]{Renato A. Krohling}
\ead{krohling.renato@gmail.com}

\author[d]{Jos{\'e} A. Ventura}

\address[a]{Federal University of Esp\'irito Santo, Brazil}
\address[b]{PPGI - Graduate Program in Computer Science}
\address[c]{Production Engineering Department}
\address[d]{Incaper, Rua Afonso Sarlo, 160, Bento Ferreira, 29052-010 Vitória, ES, Brazil}



\begin{abstract}
    Biotic stress consists of damage to plants through other living organisms. Efficient control of biotic agents such as pests and pathogens (viruses, fungi, bacteria, etc.) is closely related to the concept of agricultural sustainability. Agricultural sustainability promotes the development of new technologies that allow the reduction of environmental impacts, greater accessibility to farmers and, consequently, increase on productivity. The use of computer vision with deep learning methods allows the early and correct identification of the stress-causing agent. So, corrective measures can be applied as soon as possible to mitigate the problem. The objective of this work is to design an effective and practical system capable of identifying and estimating the stress severity caused by biotic agents on coffee leaves. The proposed approach consists of a multi-task system based on convolutional neural networks. In addition, we have explored the use of data augmentation techniques to make the system more robust and accurate. The experimental results obtained for classification as well as for severity estimation indicate that the proposed system might be a suitable tool to assist both experts and farmers in the identification and quantification of biotic stresses in coffee plantations.
\end{abstract}


\end{frontmatter}


\section{Introduction}


Plants are constantly exposed to a wide variety of biotic agents such as pests and pathogens (viruses, fungi, bacteria, etc.) and also by abiotic factors such as water deficit, heat, salinity and cold \citep{suzuki2014}. Plant diseases, caused by pathogens, lead to an impairment of the normal state of a plant, interrupting and modifying its vital functions. They can be a limiting factor in productivity and lead to significant crop losses \citep{ventura2017}. In addition, plant diseases are a threat to food security on a global scale and mainly affect small farmers whose livelihood depends on healthy crops \citep{mohanty2016}.

There are many measures that can be adopted to prevent the spread of pests and diseases in plantations. The integrated pests and diseases management in an early and controlled manner, reduces the chances of crop losses and reduce the need to use pesticides. The indiscriminate use of pesticides leads to economic losses related to their purchase and application, expenses related to medical treatment for intoxicated persons and damages caused by environmental contamination \citep{oliveira2014}. The efficient control of such biotic agents is closely related to the concept of agricultural sustainability, whose objective is to promote the development of new technologies and practices that reduce environmental impacts and be more accessible to the small farmer and increase their productivity \citep{pretty2007}. 

For the efficient control of pests and diseases,  it is important to know not only the causal agent but also the severity of the symptoms, since the diagnosis and quantification of plant stress are two equally important functions for phytopathology \citep{kranz1988}. Plant disease severity measures the percentage of the plant tissue area that is symptomatic and is important to predict yield and recommend control treatments \citep{bock2010}.


Among the sectors of the Brazilian economy, agriculture is one of the main pillars, contributing to employment generation, income and wealth for the country \citep{oliveira2014}. Brazil is the world's largest producer of coffee and this is an important crop for the country. Biotic stress such as leaf miner, rust, brown leaf spot and cercospora leaf spot, affects coffee plantations leading to the defoliation and reduction of photosynthesis, hence reducing the yield and quality of the final product \citep{ventura2017}.

Several efforts have been made using Artificial Intelligence to assist small farmers to correctly identify the diseases and pests that affect their production and the severity of the symptoms. Computer-aided diagnosis (CAD) systems allow any farmer with access to a smartphone to enjoy expert knowledge in a practical and low-cost way \citep{dehnen2016}.


The area of Artificial Intelligence research has shown an exponential growth in the few years with respect to the applications of Machine Learning, which led to the emergence of a new category of models called Deep Learning \citep{lecun2015}. Deep Learning methods have shown a clear superiority over traditional Machine Learning approaches in solving most problems. Among the methods of Deep Learning, Convolutional Neural Networks (CNN) has shown an outstanding performance in image recognition tasks \citep{fuentes2017}. CNN automatically learns the appropriate features from the training dataset while traditional approaches are based on handcrafted features, i.e., the features are calculated based on a priori knowledge of the problem. In addition, the segmentation step is intrinsic to the CNN convolutional filters, further simplifying its use.

According to \cite{barbedo2018impact} CNNs are powerful tools that deal pretty well with the problem of plant diseases. However, there are still many challenges associated with automatic diagnosis of plant diseases. In \cite{barbedo2016} the six main challenges encountered in this type of problem are addressed whereas these challenges were separated into extrinsic and intrinsic factors. Extrinsic factors are mainly related to the image acquisition in-field, i.e.: (1) Complex backgrounds, for example, leaves, soil, stones, etc. (2) Uncontrolled image capture conditions, e.g., variation of lighting, shadows, reflections and blur. Intrinsic factors: (3) Symptoms boundary poorly defined presents a gradual fading between injury and healthy tissue. (4) Symptom variations for different stages of the disease such that the interaction between plant, disease and environment leads to variations in their visual characteristics. (5) Multiple simultaneous disorders of biotic and abiotic origin may manifest on the same plant and even on the same leaf. (6) Similarities in the symptoms appearance caused by different types of stress.

In order to address the challenges pointed out by \cite{barbedo2016} new methds need to tackle issues like performance and reliability \citep{singh2018}. Challenges (4) and (5) depend heavily on the construction of more representative and reliable datasets. According to \cite{barbedo2019}, this is the main issue encountered by deep learning approaches in automatic identification of plant diseases. The creation of these datasets is not a simple task, labeling the images is a costly process and often must be carried out by a specialist. Although there are initiatives that are using the concept of social networks to accelerate this process, this practice can lead to unreliable data \citep{barbedo2018factors}.

The largest public dataset currently was developed by \cite{hughes2015} called PlantVillage and contains over than $54$ thousand images of leaves and $38$ classes, among them, healthy and diseased leaves of different species. Several studies were carried out using the PlantVillage dataset \citep{mohanty2016,too2019,kaya2019}, most of these studies presented results greater than $99\%$ of accuracy. In addition, it is common in the literature works that developed their own dataset applied to a specific type of crop such as \cite{fuentes2017} that collected images of tomato leaves using conventional cameras. \cite{johannes2017} developed a system capable of identifying diseases in photos of wheat leaves obtained by smartphones. \cite{liu2017} created a dataset of apple leaves and proposed a new architecture based on AlexNet for the recognition of diseases and \cite{ma2018} developed a dataset with images of common disease symptoms affecting cucumber leaves.

Although the problem of plant leaf disease has been addressed in several studies, few have focused on developing systems capable of estimating stress severity. \cite{wang2017} proposed the use of Convolutional Networks to estimate the severity of plant diseases. The images of apple leaves of the dataset affected by black rot were labeled in four degrees of severity. The experimental results showed an accuracy greater than $90\%$.

Typically in Machine Learning, we focus on solving a single problem, such as diagnosing leaf diseases or estimating the stress severity. To do this, usually a model or a set of models are trained to solve specific tasks. On the other hand, there is the concept of multi-Task Learning (MTL), which consists of training a system capable of solving multiple tasks using a shared architecture. MTL leads the model to prefer representations that serve for both tasks, in such a way that the initial layers of the network will learn joint generalized representations allowing a better generalization ability of the model, preventing overfitting \citep{ruder2017}.

\cite{liang2019} proposed a multitasking CAD system able to diagnose diseases, recognizing the plant species and estimating the severity of diseases, called PD$^2$SE-Net. The PlantVillage dataset was used to perform the experiments. The estimation of stress severity consisted of classifying the leaves in one out of three classes: healthy, general and serious. The results presented an overall accuracy of $91\%$ and $98\%$ for disease severity estimation and plant disease classification.

\cite{ghosal2018} developed a framework based on CNN able to identify and classify a set of 8 stresses (biotic and abiotic) affecting soybean leaves. In addition, they presented a mechanism for selecting and extracting the feature maps that best isolate visual symptoms in a similar way to that performed by humans and they had used such maps to quantify stress severity.

In a recent study, \cite{barbedo2019}  explored the use of individual lesions instead of considering the entire leaf. The lesions were manually segmented, this allowed to significantly increase the dataset and to identify multiple and different lesions  that affect the same leaf, in contrast to manual segmentation that is a laborious process mainly in leaves with many small lesions.

\cite{giuliano2019} developed a system capable of classifying the individual lesions of coffee leaves whereas the segmentation of the lesions was performed using a threshold-based method. The lesions were classified into one out of two classes, leaf miner and rust, using handcrafted features. With the lesion segmentation results it was possible to quantify the severity and to identify each lesion individually. Although the work presented good results, the segmentation with this type of approach does not take into account the positional relation between the pixels, being easily affected by factors such as illumination variation and specular reflection.


Due to the limitations of the presented works and the conventional techniques of image processing, in this work, we have significantly extended the work of \cite{giuliano2019} in the following aspects: 1) It was built a larger and more representative image dataset of  healthy and diseased coffee leaves which will be made publicly available; 2) The training and evaluation of Convolutional Neural Networks was introduced to form a more robust system for automatic classification and severity estimation of coffee leaf biotic stress.

The remainder of this paper is organized as follows. Section 2 is devoted to describing the details of the dataset and explains the proposed system architecture. Section 3 presents the experimental results and discusses our findings. Section 4 draws some conclusions and suggests directions for future work.

\section{Materials and methods}

\subsection{Image dataset}

The dataset developed for this work contains images of coffee leaves affected by the main biotic stresses that affect the coffee tree. The images were obtained using different smartphones (ASUS Zenfone 2, Xiaomi Redmi 5A, Xiaomi S2, Galaxy S8 and Iphone 6S). In addition, the leaves were collected at different times of the year and in different regions of the state of Esp\'irito Santo, Brazil. The photos were taken from the abaxial (lower) side of the leaves under partially controlled conditions and placed on a white background. The acquisition of the images was done without much criterion in order to make the dataset more heterogeneous.

\begin{figure}[htbp]
\centerline{\includegraphics[width=4.0in]{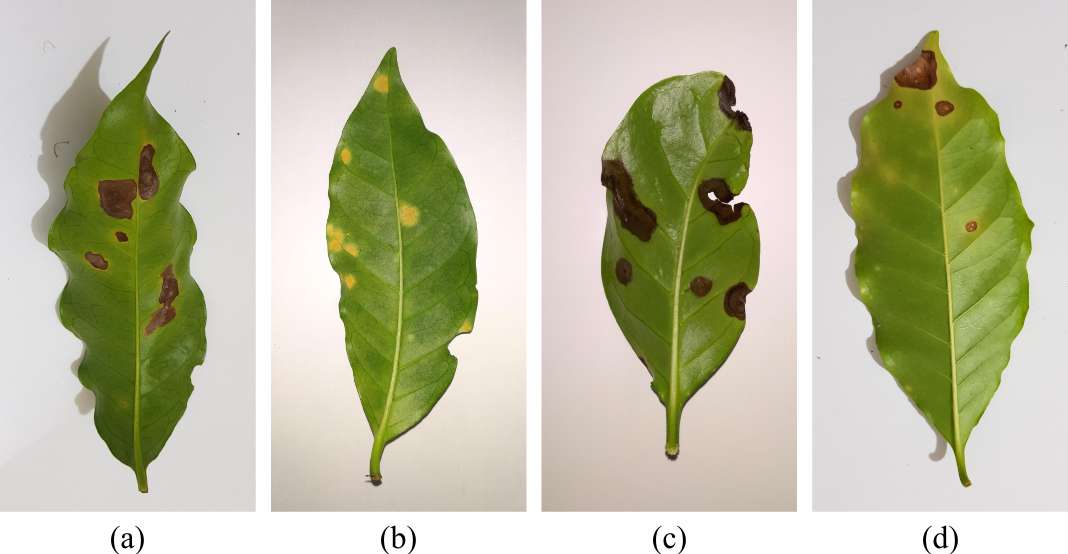}}
\caption{Examples of coffee leaves affected by different biotic stresses: leaf miner (a), rust (b), brown leaft spot (c) and cercospora leaf spot (d).}
\label{leavesexamples}
\end{figure}

A total of 1747 images of coffee leaves were collected, including healthy leaves and diseased leaves, affected by one or more types of biotic stresses. The process of recognition of biotic stresses for the labeling of images was assisted by a specialist. The dataset contains the following stresses: leaf miner, rust, brown leaf spot and cercospora leaf spot. Fig. \ref{leavesexamples} shows some examples of images contained in the dataset.

From the obtained photos were generated two datasets. A dataset with the original images of the entire leaves and a second one containing only symptoms images. Details of each dataset are described in the following.

Leaf dataset: It consists of the original images of the entire leaves. The images were labeled in relation to the predominant biotic stress of each leaf and its severity. A total of $372$ images showed leaves with more than one kind of stress, among them, $62$ leaves presented stresses with similar severity, making it impossible to visually distinguish which stress predominates. Therefore, these $62$ images were not used on this dataset. Stress severity was calculated using the symptom and leaf segmentation mask using automatic image processing methods presented in \cite{giuliano2019}. Segmentation results were visually validated for all images. The severity value for those that had poor segmentation were discarded and these images were analyzed separately by a specialist applying visual estimation methods. For certain severity ranges, labels were assigned as follows: healthy ($< 0.1\%$), very low ($0.1\% - 5\%$), low ($5.1\% - 10\%$), high ($10.1\% - 15\%$) and very high ($> 15\%$).

\begin{figure}[htbp]
\centerline{\includegraphics[width=3.5in]{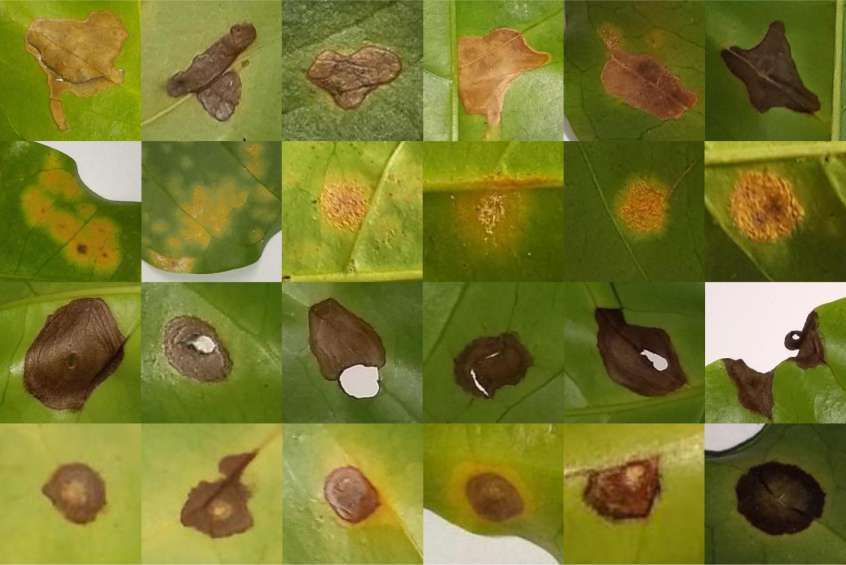}}
\caption{Examples of symptom images.}
\label{symptomsexamples}
\end{figure}

Symptom dataset: This dataset was created by cropping the isolated symptoms from the original images in a way that only a single stress was present in each image. A total of $2147$ symptom images were cropped. In addition to our images, $575$ images made available by \cite{barbedo2019} were added to our dataset, accounting for $2722$ symptom images. Fig. \ref{symptomsexamples} presents some of the symptoms extracted from the original images.

Table \ref{datasetinfo} presents the number of images for each stress/leaf and severity/leaf pair of the above datasets.

\begin{table}[htbp]
\caption{Datasets details}
\begin{center}
\begin{tabular}{ccc|cc}
\hline
\textbf{\textit{Biotic stress}} & \textbf{\textit{Leaf dataset}} & \textbf{\textit{Symptom dataset}} & \textbf{\textit{Severity}} & \textbf{\textit{Leaf dataset}}\\
\hline
Healthy & 272 & 256 & Healthy & 272 \\
Leaf miner & 387 & 593 & Very low & 924 \\
Rust & 531 & 991 & Low & 332 \\
Brown leaf spot & 348 & 504 & High & 101 \\
Cercospora leaf spot & 147 & 378 & Very high & 56 \\
\hline
Total & 1685 & 2722 & Total & 1685 \\
\hline
\end{tabular}
\label{datasetinfo}
\end{center}
\end{table}

\subsection{Data augmentation}

Complex models tend to suffer overfitting when trained with small datasets. To overcome this problem, data augmentation techniques are used to generate synthetic samples of the training data in order to increase the generalization ability of the model. In this work we used standard techniques in the literature \citep{inoue2018}, we call it standard augmentation, and we explored the use of a new approach that aims to combine two randomly selected images generating a new one. Details of the used techniques are presented in the following.

\begin{itemize}
    \item Standard augmentation: The use of standard data augmentation techniques are quite common in most works in the literature and present good performance gains. Although there are several ways to generate new images from the training samples, we focus only on the most common ones, being: horizontal and vertical mirroring, rotation and color variation (brightness, contrast and saturation). Standard augmentation were applied to all experiments.
    
    \item Mixup: This approach was proposed by \cite{mixup2017} consists of a simple linear combination of two images and their labels. Therefore, a new image is generated by $\{\tilde{x},\tilde{y}\} = \lambda \{x_i,y_i\} + (1 - \lambda)\{x_j,y_j\}$ where $\lambda \in [0,1]$ is a random value generated by the beta probability distribution, $x$ represents the set of images and $y$ the set of labels encoded in the one-hot format.
\end{itemize}

\subsection{Deep learning architectures}

Several CNN architectures are being proposed each year, each with its own particular characteristics. Despite the differences, they all share the same goal that is to increase accuracy and reduce the models complexity. Some architectures provide great performances on a wide range of applications. We selected some of the most common architectures used in the classification problem of plant diseases. Some characteristics of the used architectures are presented in the Table \ref{tabcnns}.

\begin{table}[htbp]
\caption{Properties of the CNN architectures used in this work.}
\begin{center}
\begin{tabular}{ccc}
\hline
CNN architecture & Parameters (M) & Layers \\
\hline
AlexNet & 61 & 8 \\
GoogLeNet & 6.9 & 22 \\
VGG19 & 138 & 19 \\
ResNet50 & 25 & 50 \\
\hline
\end{tabular}
\label{tabcnns}
\end{center}
\end{table}

\begin{figure}[htbp]
\centerline{\includegraphics[width=4in]{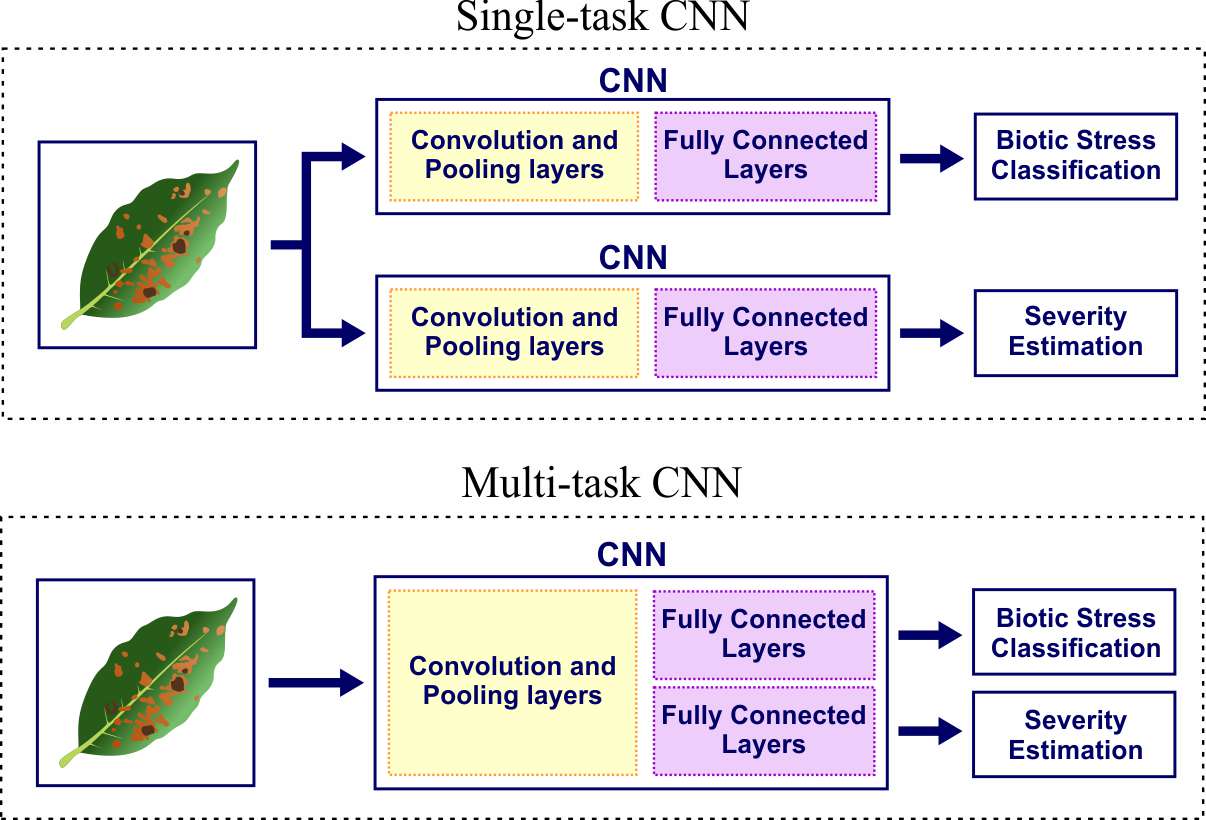}}
\caption{Framework used for single-task and multi-task learning.}
\label{multitasknet}
\end{figure}

The Leaf dataset has two tasks, biotic stress classification and severity estimation, since these tasks are closely related (problems of the same domain) is proposed in this work the adaptation of CNN networks making them multi-task systems. Therefore, the architectures were modified by the addition of a new fully connected layer in parallel with the existing one. In this way, the classification blocks are individualized but the convolutional layers are shared, i.e., the model will learn joint features useful in classifying both problems. Fig. \ref{multitasknet} presents the basic structure of systems with single-task and multi-task learning.

One can notice that multi-task learning reuses most of the network architecture. This simplification makes learning substantially faster since only a single model needs to be trained.

\section{Experiments}

Experiments were conducted in order to evaluate how appropriate the models are for the problems of biotic stress classification and severity estimation. Details of the experimental setup and the results obtained are presented in the following subsections. The source-code and the dataset is available at: 
\hyperlink{https://github.com/esgario/lara2018}{https://github.com/esgario/lara2018}

\subsection{Experimental setup}

\begin{figure}[htbp]
\centerline{\includegraphics[width=4.5in]{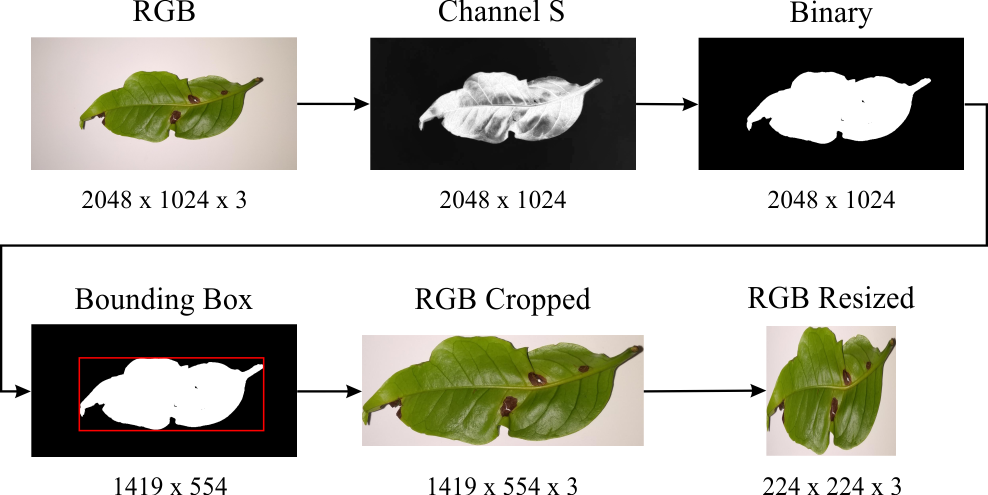}}
\caption{Cropping and resizing leaf images.}
\label{crop}
\end{figure}

To meet the input size requirements of CNN networks, images need to be resized to $224$x$224$x$3$. Due to the size of the images, the network may find difficult to capture relevant characteristics in the dataset of leaves with very small symptoms. To alleviate this problem, a traditional thresholding method based on fixed threshold was used in the $S$ channel of the $HSV$ color space, in such a way that the leaf segmentation allows us to fit a bounding box for the removal of excesses that dont belong to the region of interest. Therefore, the images were cropped and resized allowing the leaf to cover much of the image. Fig. \ref{crop} shows the steps performed to crop and resize the leaves.

For the realization of all the experiments were used the proportions $70$-$15$-$15$ for the training, validation and test datasets, respectively. Data augmentation techniques were applied online during training. Each new batch of images produces new images that are inputed into the CNN network and the network weights are adjusted until the network learns the most relevant discriminative features for a given problem. To make the training more efficient, transfer learning technique was applied. The training was performed by adapting pre-trained networks with the ImageNet dataset. The models were trained end-to-end, without freezing the training of any layer. The hyper-parameters used during the models training are presented in the Table \ref{tabparam}.

\begin{table}[htbp]
\caption{CNN training hyper-parameters.}
\begin{center}
\begin{tabular}{cc}
\hline
Parameter & Value \\
\hline
Optimizer & Stochastic Gradient Descent \\
Loss function & Cross-Entropy \\
Epochs & 100 \\
Batch size & 32 \\
Learning rate* & 0.01 \\
Momentum & 0.9 \\
Weight decay & 0.0005 \\
\hline
\multicolumn{2}{c}{\footnotesize *Decreases by a factor of 1/2 or 1/5 every 20 epochs, alternately.}
\end{tabular}
\label{tabparam}
\end{center}
\end{table}

During the training of the networks, the states (set of weights) in which the models presented the lowest loss value for the validation set were saved. The saved models were then evaluated with the test dataset and the results were computed in terms of Accuracy (acc), Precision (prc) and Recall (rec). All experiments were performed using PyTorch, a machine learning open source library. The models were trained using an NVIDIA GeForce GTX 1060 with CUDA 10.0.


\subsection{Leaf dataset results}

The obtained results for the Leaf dataset are presented in the Table \ref{results_leaf}. The results were divided by task (biotic stress and severity), CNN architecture (AlexNet, GoogLeNet, VGG16 and ResNet50) and learning approach (single-task and multi-task). The best results are presented in bold.

    \begin{table}[htbp]
        \caption{Test results obtained with different architectures for the Leaf dataset.}
        \begin{center}
        \begin{tabular}{c|c|ccc|ccc}
        \hline
        \multirow{2}{*}{Task} & \multirow{2}{*}{Architecture} & \multicolumn{3}{c|}{Single-task} & \multicolumn{3}{c}{Multi-task} \\
        & & acc(\%) & prec(\%) & rec(\%) & acc(\%) & prec(\%) & rec(\%) \\
        \hline
        & AlexNet & 92.46 & 89.37 & 90.02 & 91.67 & 88.14 & 88.67 \\
        & GoogLeNet & 91.67 & 88.33 & 89.38 & 94.05 & 92.56 & 90.87\\
        Biotic & VGG16 & 95.47 & \textbf{96.73} & 91.94 & 94.44 & 94.62 & 89.55\\
        Stress & ResNet50 & \textbf{95.63} & 94.12 & \textbf{92.7} & \textbf{95.24} & \textbf{95.29} & \textbf{91.14} \\
        & ResNet50* & \textbf{95.63} & 95.79 & 92.32 & 94.84 & 92.80 & 91.48 \\
        \cline{2-8}
        & Average & 94.17 & 92.87 & 91.27 & 94.05 & 92.69 & 90.34 \\
        \hline
        & AlexNet & 84.13 & 74.23 & 72.86 & \textbf{86.9} & 80.88 & 77.57 \\
        & GoogLeNet & 82.94 & 74.27 & 73.76  & 82.94 & 75.39 & 73.76 \\
        Severity & VGG16 & \textbf{86.51} & \textbf{82.49} & \textbf{80.89} & 86.51 & 79.50 & 76.31 \\
        & ResNet50 & 84.13 & 81.66 & 78.9 & 86.51 & \textbf{82.38} & \textbf{80.9} \\
        & ResNet50* & 78.57 & 69.33 & 67.42 & 80.95 & 78.17 & 68.97 \\
        \cline{2-8}
        & Average & 83.26 & 76.40 & 74.77 & 84.76 & 79.26 & 75.50 \\
        \hline
        \multicolumn{8}{c}{\footnotesize *Model trained with standard augmentation and mixup.}
        \end{tabular}
        \label{results_leaf}
        \end{center}
    \end{table}
    
    The results show that the system obtained on average an accuracy of $94.05\%$ for biotic stress classification and $84.76\%$ for severity estimation using multi-task learning. With the single-task learning, the obtained results were $94.17\%$ and $83.26\%$. By comparing the individual results of each architecture, we notice that AlexNet, GoogLeNet and ResNet50 have been benefited from multi-task learning. The addition of mixup in ResNet50 training presented similar results for the biotic stress classification problem, however, it worsened when applied to the severity estimation. The performance of mixup is discussed in the following subsections. In addition to the performance results of the models, we present in Table \ref{time} the average time in minutes for the training of each architecture.
    
    \begin{table}[htbp]
        \caption{Average training time for one epoch for the architectures investigated.}
        \begin{center}
        \begin{tabular}{cc}
        \hline
        Architecture & Time(sec) \\
        \hline
        AlexNet & 7.32 \\
        GoogLeNet & 11.76 \\
        ResNet50 & 21.90 \\
        VGG16 & 36.48 \\
        \hline
        \end{tabular}
        \label{time}
        \end{center}
    \end{table}
    
\begin{figure}[htbp]
        \centerline{\includegraphics[width=5in]{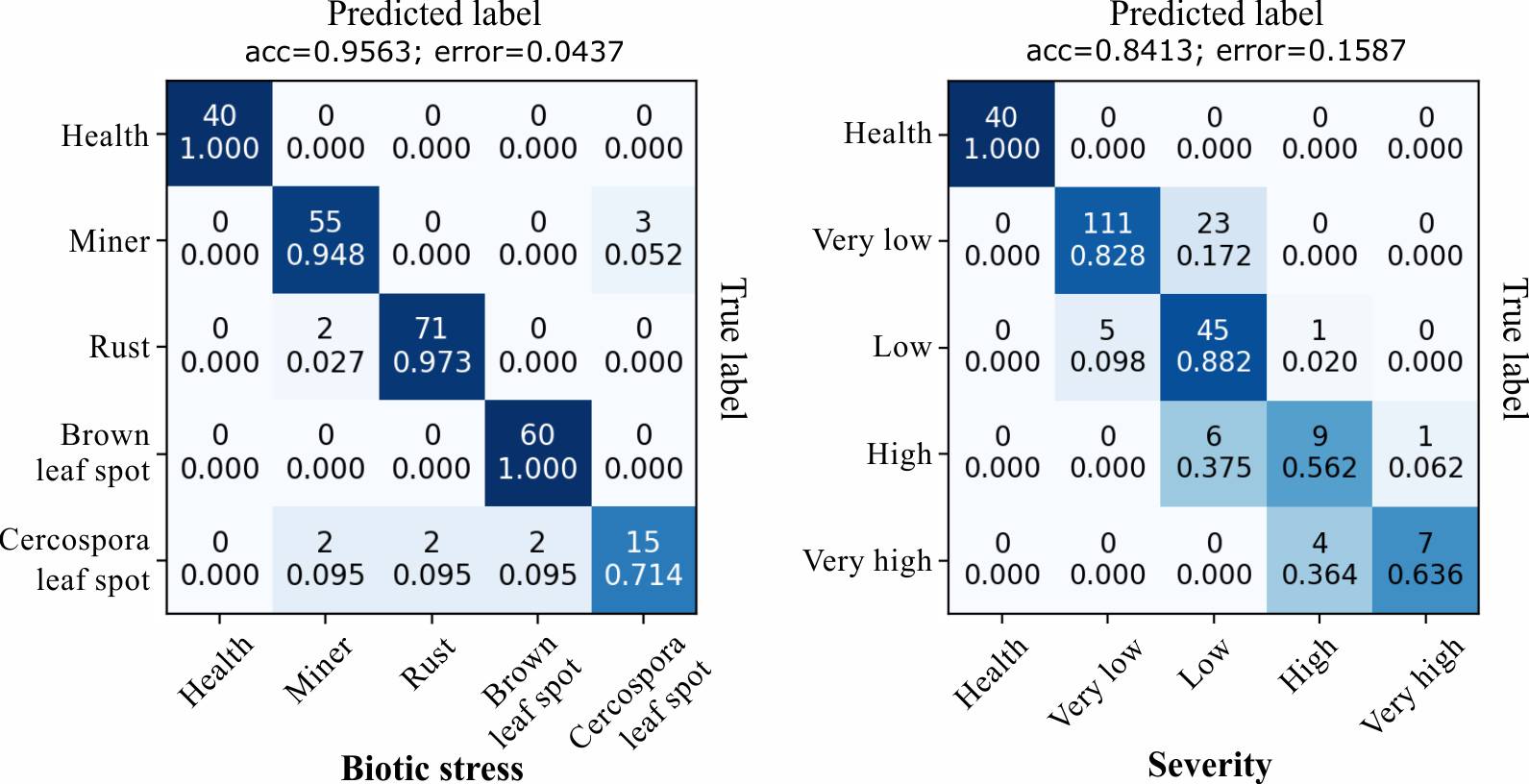}}
            \caption{Confusion matrix for the ResNet50 prediction results using the Leaf Dataset.}
        \label{cm_leaf}
    \end{figure}

    Among the used architectures ResNet50 presented a tradeoff between computational cost, performance and reliability superior to the competitors. Fig. \ref{cm_leaf} presents the confusion matrices associated with the prediction results obtained with ResNet50. The classification results of biotic stress were consistent for most of the stresses except cercospora leaf spot which presented a considerable amount of classification errors. This result corroborates with the experiments carried out in \cite{barbedo2019} whose class with the largest number of samples misclassified was also the cercospora leaf spot. These misclassifications may be associated with similarity with other diseases and also with the dataset imbalance.
    
    The severity estimation with the ResNet50 obtained an accuracy of $84.13\%$. This problem is clearly more challenging compared to biotic stress classification. The model presented no difficulty in separating healthy leaves from diseased, however, due to imbalance of the data set, the classification of leaves with high and very high severity presented the highest misclassification rates. Although the model has made a considerable amount of mistakes, we notice by the confusion matrix that these errors are located close to the main diagonal. Therefore, we can say that the model did not make errors considered grotesque. Due to the ordinal nature of severity labels, estimating the stress severity as high rather than very high, for example, may be considered a minor error for this problem. 
    
\subsection{Symptom dataset results}

    The experiments performed with the Symptom dataset obtained on average an accuracy of $96.63\%$ as presented in Table \ref{results_symptom}. As expected, the classification results only of the symptoms are superior as compared to the classification using images of the entire leaves. This occurs because the symptoms images have a well delimited region of interest which leads to the elimination of spurious elements. So, the network can then focus on and visualize a greater level of symptoms details. The ResNet50 architecture once again provided the best results, unlike VGG16 which presented good results for the Leaf dataset but poor performance for the Symptom dataset.

    \begin{table}[htbp]
        \caption{Test results obtained with different architectures for the Symptom dataset.}
        \begin{center}
        \begin{tabular}{c|ccc}
        \hline
        \multirow{2}{*}{Architecture} & \multicolumn{3}{c}{Biotic stress} \\
        & acc(\%) & prec(\%) & rec(\%) \\
        \hline
        AlexNet & 96.58 & 96.12 & 96.59 \\
        GoogLeNet & 96.82 & 96.56 & 96.64 \\
        VGG16 & 95.60 & 95.02 & 95.31 \\
        ResNet50 & \textbf{97.07} & \textbf{96.85} & 96.69 \\
        ResNet50* & \textbf{97.07} & \textbf{96.85} & \textbf{96.99} \\
        \hline
        Average & 96.63 & 96.28 & 96.44 \\
        \hline
        \multicolumn{4}{c}{\footnotesize *Trained model with Standard aug. and mixup.}
        \end{tabular}
        \label{results_symptom}
        \end{center}
    \end{table}
    
     \begin{figure}[htbp]
        \centerline{\includegraphics[width=5in]{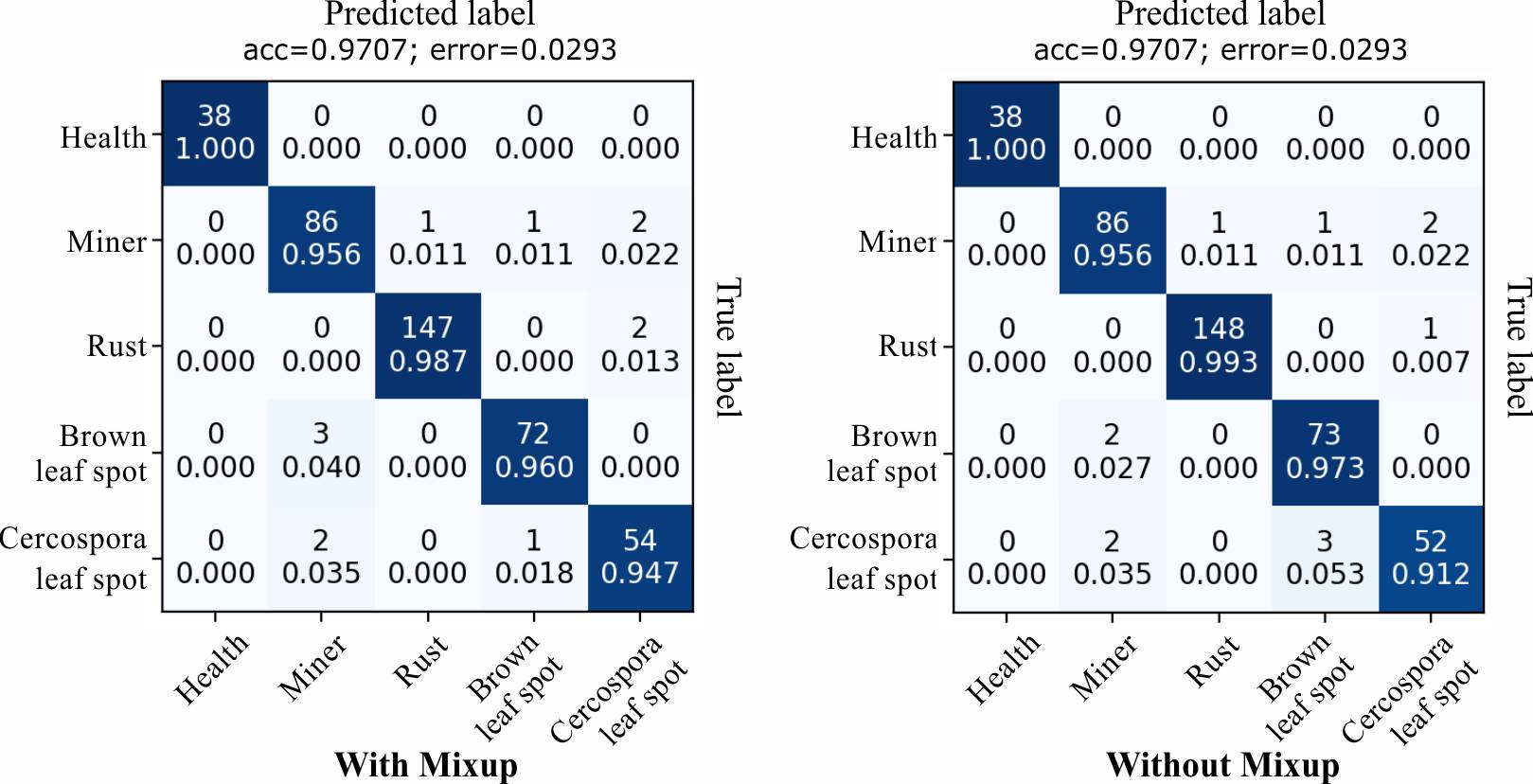}}
        \caption{Confusion matrix for the ResNet50 prediction results using the Symptom Dataset.}
        \label{cm_symptom}
    \end{figure}

The addition of mixup to ResNet50 training presented a small gain of recall compared to the model trained only with standard augmentations. Although performance can be considered the same, it was realized that the use of this technique reduces the risk of overfitting during the training stage, due to the great diversity of new images that are generated throughout the learning process. Fig \ref{cm_symptom} presents the confusion matrices with the prediction results obtained by ResNet50 trained with and without mixup. The higher recall value reflects better balanced results between classes. Therefore, the mixup was able to deal better with imbalanced data.

\subsection{Visualization}

\begin{figure}[htbp]
\centerline{\includegraphics[width=6.5in]{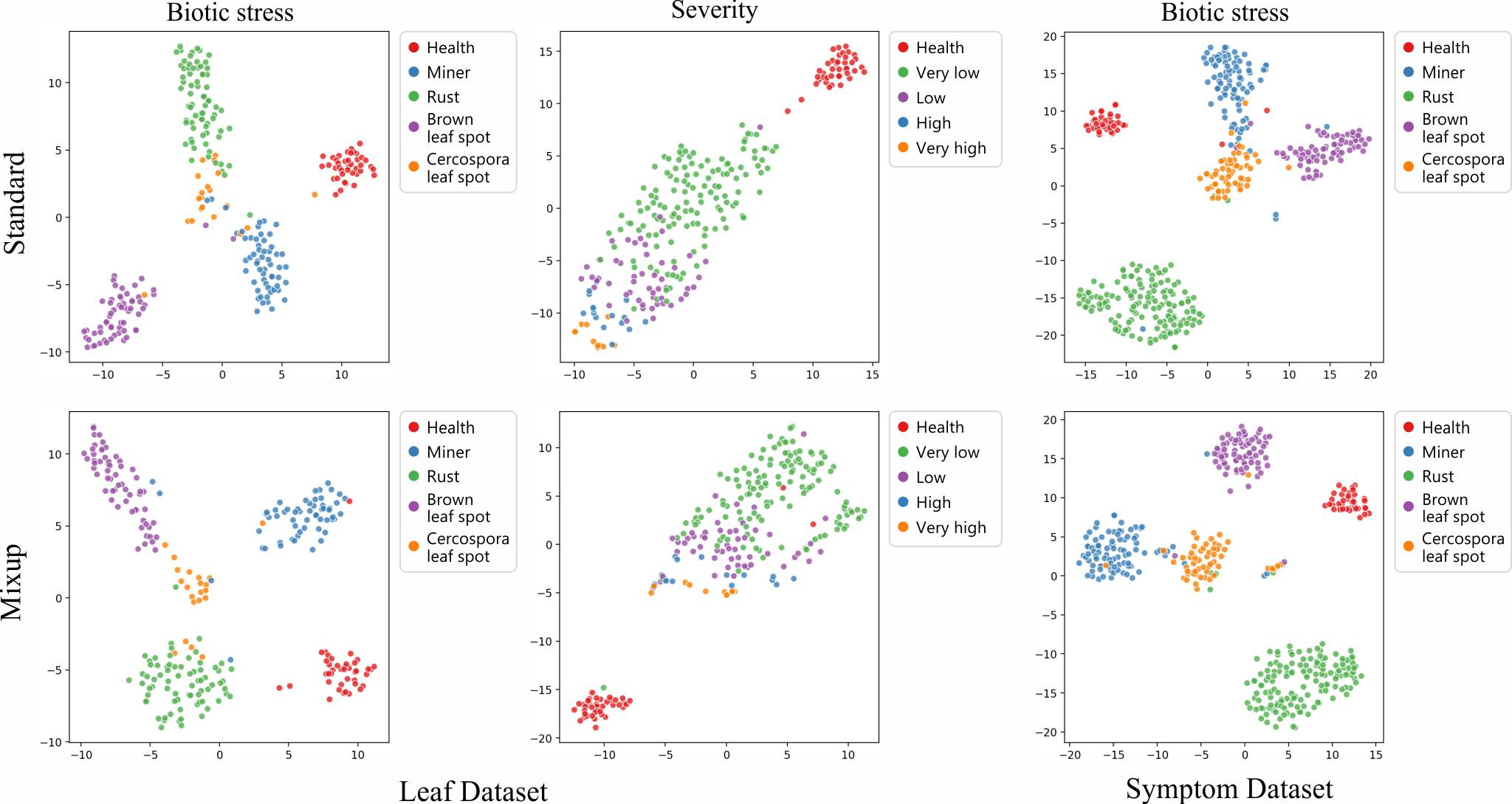}}
\caption{Visualization of the features spatial distribution extracted by ResNet50 from the test samples using t-SNE.}
\label{featuresvisualization}
\end{figure}

The t-Distributed Stochastic Neighbor Embedding (t-SNE) \cite{maaten2008} technique was used in the visualization of the high-dimensional features extracted by CNN as depicted in Fig. \ref{featuresvisualization}. In total, $2048$ features were extracted from each test sample after the last ResNet50 convolutional layer. These features were reduced to two dimensions using t-SNE.

By visualizing the extracted features for the problem of biotic stress, we note that the cercospora leaf spot presented the highest overlap with other classes, showing the difficulty of the model to discriminate this class. Nevertheless, the model presented well defined clusters of each class, showing that the model has a high generalization ability. In addition, the severity estimation with standard augmentation showed a very linear relationship between the labels and their severity range.

The features learned with mixup presented a smaller within-class variance when compared to standard augmentation. Especially for the Symptom dataset where it is clear how the features learned by the mixup have a more spherical format. Although the mixup  has presented an appropriate behavior to the classification problem of biotic stress the technique worsened the classification results when applied to the severity estimation. The analysis of the results suggests that due to the ordinal nature of the severity labels, the combination of images with the mixup produced images with characteristics of median severity, which biased the network to learn more intermediate labels than extreme labels, causing a higher overlapping of the features.

\section{Conclusion}

This work presented different approaches using Deep Learning for the problems of biotic stress classification and severity estimation of the most important coffee diseases and pests through leaf images. For the accomplishment of the experiments a new dataset of coffee leaves images was developed. Different CNN architectures were used in experiments, whereas the trained network ResNet50 was the one that obtained the best results. It was verified that the multi-task learning can make the model more effective to solve tasks of the same domain in both performance and computational cost. One limitation of this work is related to the low representativity of the dataset that covers only the main biotic stresses that affect coffee trees. However, the results using Deep Learning techniques have proven consistent, increasing the dataset with new kind of stresses and a higher number of images might help alleviate this problem. Since an App has been developed in our lab, we are embedding this new technology to be used in practice. The first tests have been carried out with promising results and will be reported in near future.

\section*{Acknowledgments}
J.G.M. Esgario and R.A. Krohling thank the funding support provided by Google Latin America Research Award (LARA 2018-2019).

\section*{References}

\bibliographystyle{model5-names.bst}\biboptions{authoryear}

\end{document}